# Bach Style Music Authoring System based on Deep Learning

*Minghe Kong, Lican Huang*

Informatics college, Zhejiang Sci-Tech University，Hangzhou ,China,310018

**Abstract:** With the continuous improvement in various aspects in the field of artificial intelligence, the momentum of artificial intelligence with deep learning capabilities into the field of music is coming. The research purpose of this paper is to design a Bach style music authoring system based on deep learning. We use a LSTM neural network to train serialized and standardized music feature data. By repeated experiments, we find the optimal LSTM model which can generate imitation of Bach music. Finally the generated music is comprehensively evaluated in the form of online audition and Turing test. The repertoires which the music generation system constructed in this article are very close to the style of Bach's original music, and it is relatively difficult for ordinary people to distinguish the musics Bach authored and AI created.

## 1  INTRODUCTION

With the development of digital media technology, traditional music has grown rapidly, and the popularity and dissemination of music have reached the peak of history. Johann Sebastian Bach (1685-1750) has rich musical genres and spread widely. He has achieved gorgeous and unrestrained Baroque music without losing dignity, which is known as the "father of western music" [1]. At the same time, Bach's music has obvious musical theory characteristics and strong regularity of interval and rhythm, which is conducive to neural network to extract the characteristics. In algorithmic composition and artificial intelligence composition, the research core of researchers is how to make computers understand and find out the probability distribution and digital logic behind complex music.

Therefore, we use LSTM (Long Short Term Memory) deep neural network model to learn Bach's classical repertoire including trio, French Suite, ancient suite, Brandenburg suite and so on. As far as we know, this is the first study to apply neural network to Bach style music creation. The training rules are as follows: predict the value of the next note and duration through the note and duration sequence of a given length, compare it with the expected value in the training set, and modify LSTM weights to make less error through loss function and to get better results. The training of note sequence uses Bi-LSTM layer to consider the forward and backward information to make the prediction more accurate.

## 2  Related Work

Iannis Xenakis tried to compose music with Hidden Markov Model (HMM) in 1958. For example, Generative.fm[2] can continuously generate new music sentences with excellent effect through Markov chain and few music samples. However, HMM can only generate and combine existing music clips, and the music is lack of diversity.

Musegan [3] proposed in 2018 can generate small pieces of multi track music by using the method of Generative Adversarial Networks (GAN)[4], and there is no ambiguity problem in the generated samples. However, the training of GAN is difficult, and its training process is often difficult to converge, and there will be frequent vibrations. And the experimental results are random, and difficult to reproduce. After the model converges, it is also prone to pattern collapse, and many training skills are needed to avoid pattern collapse. If we use the method of GAN to generate multi track music with more reasonable length, the scale of the network will become too large and the training cost will rise greatly.

In 2019, OpenAI launched jukebox [5], which can directly learn music sound wave data and sample music sound waves' model . This model is trained by using combining Vector Quantized Variational Auto-encoder and WaveNet [6], with compressing music into discrete code and reconstructing music. However, this technology is not mature. It takes nine hours to generate a one minute sound file using this model [5].

It is very popular to synthesize music score based on recurrent neural network model. If you want to pursue better results, the notes output from the network should not only consider the pitch and chord, but also consider the duration of the notes. Such prediction is difficult for the LSTM model. In this paper, the note generation network is constructed by using Bi-LSTM layer, and the smaller duration generation network is constructed by using the traditional LSTM model. Combined with the normalization of duration sequence, the network complexity and raining cost is greatly reduced without losing final automatic composition effect.

## 3 Music Data Expression and Processing

### 3.1 Music Data acquisition

Combined with relevant literature, considering multimodal analysis recognition method [7], $88 \times m$ filter array method [8] and sequence modeling method [9], this paper selects MIDI format file as music data source and music21 as analysis library for processing music to obtain music data. Use the converter.parse method to parse the music file to get the music stream object. Traverse the stream to obtain each note or chord element object, and further obtain the distribution of each note in the song in detail.

### 3.2 Music Data Processing

The huge amount of data in the training set can ensure that the training results achieve the expected goals, but it will also bring the problems of difficult data fitting and long training time, especially when a large number of data sets involve chords composed of multiple notes.

After conceiving and trying the virtual piano key method and chord fusion method, the chord note separation method is the final method chosen. Separate the individual notes from the music without damaging the overall characteristics of the chord, and finally form a note Chord Dictionary. The $10^6$ magnitude music data is represented by the $10^3$ magnitude dictionary, which makes the subsequent processing easier. However, because the music data is processed in the form of note sequence table, it also causes the problem that notes and chords seize the same track. Therefore, we should introduce the important concept of duration.

The duration represents the duration of the note or chord. The introduction of the duration dimension breaks the limitation of the sequence table, making the performance of Single track music similar to that of multi-voice ensemble. In this paper, all durations are normalized by quarter-length. The most common three kinds of durations are Eighth notes, quarter notes, and half notes. Generate a duration dictionary with only three key value pairs, normalize all note durations corresponding to the dictionary, and wait for the next processing.

## 4 Deep Learning Network Model Construction

The deep neural network mainly involved in this study is long short-term memory (LSTM) [10-14] network, which can solve tasks that cannot be solved by RNN [15] learning algorithm.

4.1 Train Data set preprocess and vectorization

Through the previous processing, we get the note sequence and duration sequence. Taking the note sequence as an example, if the length of the note sequence is time t, there is Note = [$note_1$、$note_2$、……$note_t$ ], assuming that each neural network input is n notes, the network output $note_{n+1}$ is obtained. A batch uses m sequences to form the input note set X and the output note set R:

$$X = \begin{bmatrix} note_1 & note_2 & note_3 & & note_n \\ note_2 & note_3 & note_4 & \cdots & note_{n+1} \\ note_3 & note_4 & note_5 & & note_{n+2} \\ & \vdots & & \ddots & \vdots \\ note_m & note_{m+1} & note_{m+2} & \cdots & note_{m+n} \end{bmatrix} \quad R = \begin{bmatrix} note_{n+1} \\ note_{n+2} \\ note_{n+3} \\ \vdots \\ note_{n+m+1} \end{bmatrix}$$

Figure 1 Input and OutputNote Sets

Similar to the note network, the input value of the duration neural network is set as the splicing of the duration matrix and the note matrix, and the output value is the duration matrix, that is, the music is generated first, and then the rhythm of the music is determined according to the music melody.

4.2 Music segmentation algorithm

Since all notes are arranged into a note sequence, when the end position k of the song is between $note_1$ and $note_n$, it is bound to use the data of the previous song to try to train the output of the current song. This abnormal input-output relationship will not only affect the efficiency of network training, but also destroy the existing good connection and make the acquisition mode unstable. When the intercept sentence length is small, it is not obvious, but the influence of this wrong segmentation method will gradually increase with the increase of the intercept sentence length.

$$I = \frac{sequence\_length}{\frac{\sum_{i=1}^{n} song\_length_i}{n}} \quad (1)$$

The music segmentation algorithm adds an end flag '&' to the note dictionary to mark the end of the song. At the same time, it can help the generator jump to the beginning of the next song when training to the end of the song.

4.3 Neural Network Structure implementation and selection of parameters

The following neural networks are finally selected through multiple experimental training and comparison of the generated results. The specific network architecture and network implementation details are shown in table 1:

Table 1 model1 parameters of each layer

| Neural Network Layer | parameters |
|---|---|
| LSTM | 512 |
| Dropout | Keep=0.7 |
| BidirectionalLSTM | 256 |
| Dropout | Keep=0.7 |
| BatchNorm | / |
| Dense | Number of note dictionaries |
| Activation | Softmax |

  The above table shows the note generation network architecture in this study. The whole network has 7 layers. Note generation is mainly realized by using LSTM network combined with bidirectional LSTM network. The parameters in LSTM layer are large in order to process a large amount of data. The dropout layer can prevent the network from over fitting. BatchNorm layer, so that the input of each layer of neural network remains the same distribution in the training process. The sense layer cooperates with the activation function softmax layer to form the final output of the network. RMSprop is selected as the network optimizer, which modifies the accumulation of the square sum of the gradient and an exponentially weighted moving average of AdaGrad , making it better under the non convex setting.

  The second layer uses bidirectional LSTM layer, which is a better construction method after many experiments. BidirectionalLSTM can consider forward and backward information at the same time, and make the semantic prediction results more accurate through comprehensive consideration.

Table 2 model2 parameters of each layer

| Neural Network Layer | parameters |
|---|---|
| LSTM | 512 |
| LSTM | 256 |
| Dropout | Keep=0.7 |
| Dense | Number of duration dictionaries |
| Activation | Softmax |

  In contrast, the duration generation network shown in Table 2 is much smaller, because the duration generation has a small output, so it is relatively unnecessary to consider too much audio information. Smaller networks can also greatly speed up the training speed and achieve better training results.

  In the face of massive training set data, the memory and GPU video memory cannot be accommodated. At this time, a new method needs to be used to input the data into the neural network model in batches. This article uses the model.fit_generator function provided by Keras. This generator function cooperates with the python generator to read the data stored in the hard disk into the model in batches, so as to exchange time for space and solve the problem of memory overflow.

## 5 Model Train and Music Generation

### 5.1 Model Train

After building the neural network and generator, start to use the network for training. As shown in Figure 2, the loss function of the above neural network converges rapidly in a few epochs, showing a downward trend of oscillation locally, not approaching the local minimum, and the neural network training reaches a good level.

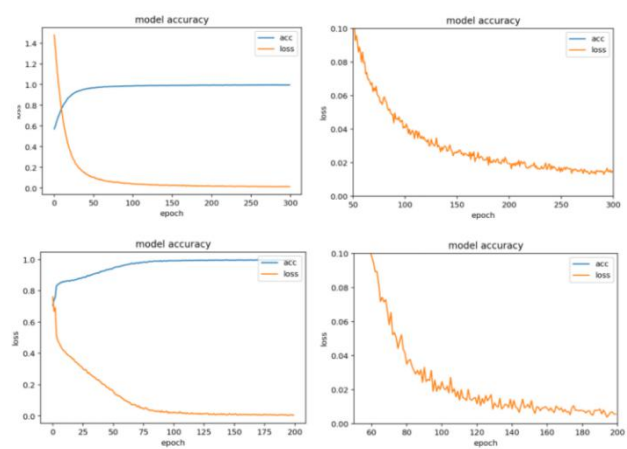

Figure 2 Model Training of loss and accuracy values

### 5.2 Music generation

When using the trained neural network for prediction generation, we need to use a random matrix with the same shape as the neural network input matrix, which is recorded as T. Through this matrix, we use model.predict to obtain the probability distribution matrix of softmax layer output, and use np.argmax function to obtain the note with the highest probability as the prediction output, Then take this prediction result as the latest bit of the input matrix to continue the next round of prediction. After several rounds, the prediction sequence is obtained. General prediction sequence length (predict_ length) is:

$$predict\_length = Length - 2 * sequence\_length \qquad (2)$$

After the generation of double sequence prediction, use music21 to decode the MIDI file to reversely edit the generated sequence into the form of stream music stream, and then package it into MIDI file. Set the note and duration values according to the sequence, and use the music score tool (MuseScore) to generate the final result, as shown in Figure 3:

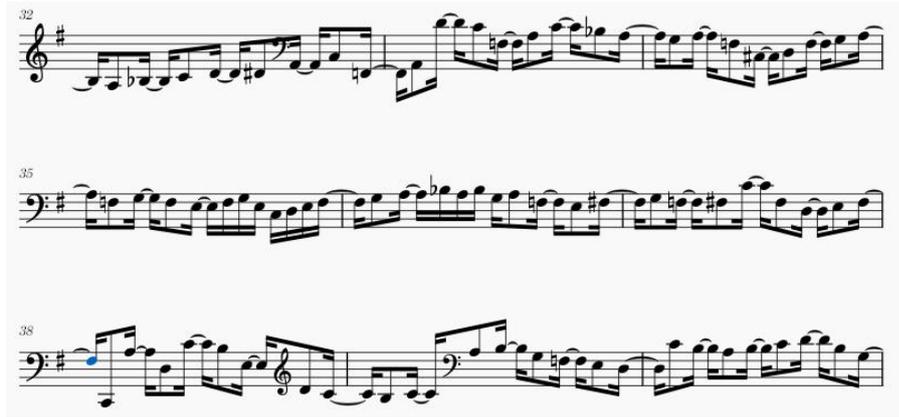
Figure 3  Music Score Generation

  The mixed generation of notes and durations can make the music more regular, rhythmic and sentence segmentation more obvious. Fig. 4 shows a music score diagram of the end of a music generated by the neural network model. Only notes are generated on the left, and notes and durations are mixed on the right. It is not difficult to see that the music generated by the mixed mode is closer to that made by human beings, and the appropriate pause makes the end of the music not abrupt and abrupt.

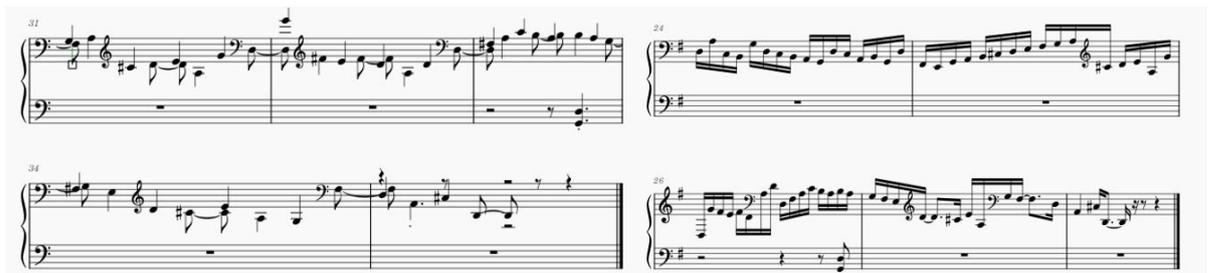
Figure 4 Comparison between  music score diagrams of the ends of  musics

5.3 Evaluation

  The effect of  music automatically generated is evaluated by online audition evaluation. After recording the three automatic composition music and randomly disturbing the order with the three original Bach musics recorded,   each song  is sampled for 20 seconds.  Online audition personnel are invited to conduct audition evaluation. The evaluation indicators are divided into three aspects: harmonious interval, smooth rhythm and beautiful melody. Each aspect is divided into five grades from low to high: very poor, poor, average, good and excellent.

  Taking the first Cello Suite No.1 in G bwv1007: I. prelude of Bach's famous repertoire in G major as the full score index, six pieces of music were evaluated. Online testers can only get the number of each track, not the composer's information. After the test, the details and scores of the six tracks are shown in Table3:

Table 3  Music Evaluation Scores

| number | Order | Author | Interval Score | rhythm Score | Melody Score | Overall Score |
|---|---|---|---|---|---|---|
| M1 | 1 | Bach | 4.0 | 4.7 | 3.8 | 4.17 |
| M2 | 2 | LSTM | 3.8 | 4.3 | 3.7 | 3.93 |
| M3 | 6 | Bach | 3.3 | 3.5 | 3.4 | 3.40 |
| M4 | 4 | LSTM | 3.8 | 4.0 | 3.8 | 3.87 |
| M5 | 5 | LSTM | 4.0 | 3.8 | 3.5 | 3.77 |
| M6 | 2 | Bach | 3.8 | 4.2 | 3.8 | 3.93 |

In the final scoring results, the music clips created by LSTM ranked second, fourth and fifth respectively. The test result on whether to distinguish the musics composed by Bach from the musics composed by artificial intelligence is shown in table 4:

Table4 Distinguishing Result

| | M1 | M2 | M3 | M4 | M5 | M6 | Not sure |
|---|---|---|---|---|---|---|---|
| Distinguished as the music composed by AI | 15.79% | 21.05% | 26.32% | 21.05% | 26.32% | 36.84% | 15.79% |

Considering that some music lovers or professionals are familiar with Bach's works, the further analysis is designed to eliminate the influence of the tracks they have heard on the second test, as shown in table 5:

Table 5 familiar Music

| | M1 | M2 | M3 | M4 | M5 | M6 | Have not heard |
|---|---|---|---|---|---|---|---|
| Having heard the music before test | 21.05% | 0.00% | 0.00% | 0.00% | 0.00% | 10.53% | 73.68% |

From the above results, it can be seen that the music generation system constructed in this paper is very close to the style of Bach's original music, and can reach a more accurate level in interval. It is difficult for ordinary people to distinguish the real creator of the music.

## 6 Conclusions

This paper constructs a music authoring system based on deep learning. The main work of this paper is as follows:

(1) Collect, analyze and extract a large number of data information of piano music, record and standardize music information in the form of note sequence and duration sequence.

(2) By trying to preprocess the data through a variety of methods, and by many experiments based on LSTM and training, finally we get a more suitable network model and automatic composition works on the basis of many experiments and training.

(3) The evaluation of music quality by using online survey shows that the music generation system constructed in this paper is very close to the style of Bach's original music, and is difficult for ordinary people to distinguish the real creator of the music.

This paper focuses on the music situation of single track and tries to use single track to simulate multiple tracks. In the future, we will study the coordination of phrases between multiple tracks; For music, besides notes and durations , intensity and emotion are important components to be considered. We will use more music theory knowledge to improve the model and generate better quality of music.